\documentclass[10pt,conference,a4paper]{IEEEtran}
\ifCLASSINFOpdf
\else
\fi

\usepackage{array}
\usepackage{booktabs}
\usepackage{graphicx}
\usepackage{subfigure}
\usepackage{multirow}
\usepackage{balance}
\usepackage{cite}

\begin{document}
%

\title{SCUT-FBP5500$^{\ddag}$: A Diverse Benchmark Dataset for Multi-Paradigm Facial Beauty Prediction}


\author{Lingyu~Liang, Luojun~Lin, Lianwen~Jin*, Duorui~Xie and Mengru~Li\\
South China University of Technology, Guangzhou 510641, China\\
E-mail:\{lianwen.jin, lianglysky\}@gmail.com\\
$^{\ddag}$Dataset Download URL: \underline{https://github.com/HCIILAB/SCUT-FBP5500-Database-Release}
}

%


%


\maketitle

\renewcommand{\thefootnote}{\fnsymbol{footnote}} 
\setcounter{footnote}{-1}

\begin{abstract}
Facial beauty prediction (FBP) is a significant visual recognition problem to make assessment of facial attractiveness that is consistent to human perception. To tackle this problem, various data-driven models, especially state-of-the-art deep learning techniques, were introduced, and benchmark dataset become one of the essential elements to achieve FBP. Previous works have formulated the recognition of facial beauty as a specific supervised learning problem of classification, regression or ranking, which indicates that FBP is intrinsically a computation problem with multiple paradigms. However, most of FBP benchmark datasets were built under specific computation constrains, which limits the performance and flexibility of the computational model trained on the dataset. In this paper, we argue that FBP is a multi-paradigm computation problem, and propose a new diverse benchmark dataset, called SCUT-FBP5500, to achieve multi-paradigm facial beauty prediction. The SCUT-FBP5500 dataset has totally 5500 frontal faces with diverse properties (male/female, Asian/Caucasian, ages) and diverse labels (face landmarks, beauty scores within [1,~5], beauty score distribution), which allows different computational models with different FBP paradigms, such as appearance-based/shape-based facial beauty classification/regression model for male/female of Asian/Caucasian. We evaluated the SCUT-FBP5500 dataset for FBP using different combinations of feature and predictor, and various deep learning methods. The results indicates the improvement of FBP and the potential applications based on the SCUT-FBP5500.\footnote{This work was supported in part by the National Natural Science Foundation of China (NSFC) under Grant~61472144 and Grant~61502176; in part by GDSTP under Grant~2015B010101004, Grant~2015B010130003, Grant~2015B010131004; in part by the National Key Research \& Development Plan of China under Grant~2016YFB1001405; and in part by Fundamental Research Funds for the Central Universities (No.2017BQ058). *Corresponding author: Lianwen Jin.} 
\end{abstract}





%
\IEEEpeerreviewmaketitle

\section{Introduction}
Assessing facial beauty seems natural for human being, but an absolute definition of facial beauty remains elusive. Recently, facial beauty prediction (FBP) have attracted ever-growing interest in the pattern recognition and machining learning communities~\cite{2011survey,zhang2016computer,laurentini2014computer}, which aims to achieve automatic human-consistent facial attractiveness assessment by a computational model. It has application potential in facial makeup synthesis/recommendation~\cite{zhang2016computer,2015liu,2011eg}, content-based image retrieval~\cite{2012ava}, aesthetic surgery~\cite{zhang2016computer}, or face beautification~\cite{2017liang,2015deep,2014liang,2008sig}.

\begin{figure}[!t]
\centering
\includegraphics[width=3.5in]{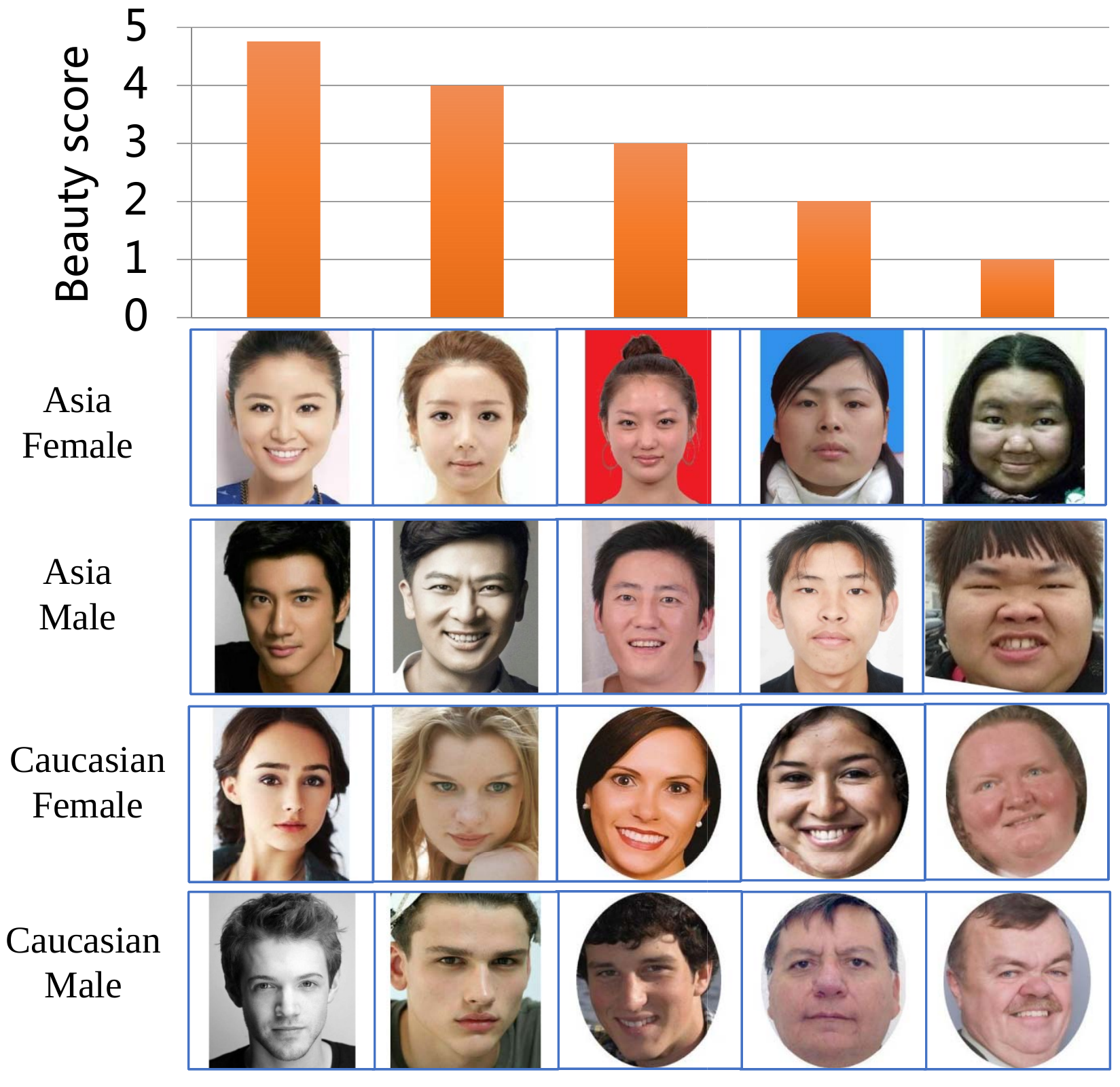}
\caption{The images with different facial properties and beauty scores from the proposed SCUT-FBP5500 benchmark dataset. The dataset download URL is shown below the title.}
\label{fig_scutfbp5500}
\end{figure}

\begin{table*}
\renewcommand{\arraystretch}{1.3}
\caption{Representative Databases for Facial Beauty Prediction}
\begin{center}
\label{table_1}
\begin{tabular}{|c||c|c|c|c|c|c|}
\hline
Database & Image Num.  & Labelers/Image& Beauty Class & Face Property & Face Landmarks & Publicly Available\\ 
\hline
\hline
Y. Eisenthal et al.~\cite{2006nc_Dror} & 184 & 28 or 18 & 7 & Caucasian Female& $\times$ & $\times$\\
\hline
F. Chen et al.~\cite{4} & 23412 & unknown & 2 & Asian Male/Female& $\surd$ & $\times$ \\
\hline
H. Gunes et al.~\cite{17} & 215 & 46 & 10 & Female & $\surd$  & $\times$ \\
\hline
J. Fan et al.~\cite{2012fan} & 432 & 30 & 7 & Generated Female& $\surd$ & $\times$ \\
\hline
M. Redi et al.~\cite{14} & 10141 & 78-549 & 10 & Multiple (Sampled from AVA~\cite{2012ava}) & $\times$ & $\surd$ \\
\hline
SCUT-FBP~\cite{2015xie} & 500 & 70 & 5 & Asian Female& $\surd$ & $\surd$ \\
\hline
\textbf{SCUT-FBP5500} & 5500 & 60 & 5 & Asian/Caucasian; Male/Female& $\surd$ & $\surd$ \\
\hline
\end{tabular}
\end{center}
\end{table*}

From the computational perspective, FBP is still a challenging problem. It is involved with the formulation of visual representation and predictor for the abstract concept of facial beauty. To tackle this problem, various data-driven models, were introduced into FBP. One line of the works follows the classic pattern recognition process, which constructs the FBP system using  the combination of the hand-crafted features and the shallow predictors. The related hand-crafted feature derived from visual recognition includes the geometric features, like the geometric ratios and landmark distances~\cite{2011survey,2009mao,2006nc_Dror,2006Kagian,2012fan,2011zhang}, and the texture features, like the Gabor-/SIFT-like features~\cite{2013wacv,2010jin,2014yan}. Then, a shallow FBP predictor
is trained by the extracted feature in a statistical manner.

Another line of works is advanced by the reviving of neural networks, especially the stat-of-the-art deep learning~\cite{2015nature}. The hierarchial structure of the deep learning model allows to build an end-to-end FBP system that automatically learns both the representation and the predictor of facial beauty simultaneously from the data. Many works
indicate that FBP based on deep learning is superior to the shallow predictors with hand-crafted facial feature~\cite{2014wang,2014gan,2010gray,xu2017facial,2017liang-icip,2015xie}.

Most of current FBP models are data-driven, which makes benchmark dataset become one of the essential elements for FBP. There have been many works of the benchmark datasets~\cite{2012fan,2014yan,17,2014wang,2010gray,4,2012ava,14} involved with FBP, but most of these datasets focus on a specific problem with specific computation constrains, as shown in Table~\ref{table_1}. Yan et al.~\cite{2014yan} regarded FBP as a ranking problem, and proposed a dataset with low-resolution images gathered from social networks. Fan et al.~\cite{2012fan} focused on the geometry analysis of FBP and proposed a dataset containing computer-generated faces with different facial proportions. The Northeast China database~\cite{4}, Shanghai database~\cite{2011zhang}, Hot-Or-Not database~\cite{white2004automatic,2010gray}, AVA database~\cite{2012ava} and re-sampled face subset of AVA database~\cite{14} are large-scale databases involved with FBP, where the Northeast China~\cite{4} and Shanghai database~\cite{2011zhang} are limited for geometric facial beauty analysis without attractiveness ratings; Hot-Or-Not database~\cite{white2004automatic,2010gray} only focuses on the appearance-based FBP; and the AVA database~\cite{2012ava,14} is originally designed for aesthetic analysis of entire images but not the facial attractiveness. In our previous work, Xie et al.~\cite{2015xie} published a SCUT-FBP benchmark dataset, which has led to many FBP models~\cite{2015xie,xu2017facial,2017liang-icip,Ren2017sense,liu2016dis}, especially the hierarchial CNN-based FBP models with the state-of-the-art deep learning~\cite{2015xie,xu2017facial,2017liang-icip,liu2016dis}. Despite the prevalent usage of the SCUT-FBP, it only contains 500 Asian Female faces, which may limit the performance of the data-demanded model for FBP. %

We find that FBP have been formulated the recognition of facial beauty as a specific supervised learning problem of classification~\cite{2014chiang,2009mao,2006nc_Dror,2014wang}, regression~\cite{2006Kagian,2012nc_mu,2012fan,2014gan,16} or ranking~\cite{2014yan,2013wacv,liu2016dis}. It indicates that FBP is intrinsically a computation problem with multiple paradigms. Previous databases built under specific computation constrains would limit the performance and flexibility of the computational model trained on the dataset, and it is difficult to compare different models derived from the dataset with specific computation paradigm. Therefore, this paper argues that FBP is a multi-paradigm computation problem, and proposes a new diverse benchmark dataset, called SCUT-FBP5500, to achieve multi-paradigm facial beauty prediction.

The SCUT-FBP5500 dataset has totally 5500 frontal faces with diverse properties (male/female, Asian/Caucasian, ages) and diverse labels (face landmark, beauty score, beauty score distribution), which allows different computational model with different FBP paradigms, such as appearance-based/shape-based facial beauty classification/regression/ranking model for male/female with Asian/Caucasian. Furthermore, the diverse faces with beauty scores gathered from 60 different labelers can lead to many interesting research, such as cross-culture facial beauty analysis, personalized FBP~\cite{19}, or automatic face beautification~\cite{2017liang,2015deep,2014liang,2008sig}. Both shallow prediction model with hand-crafted feature and the state-of-the-art deep learning models were evaluated on the dataset, and the results indicates the improvement of FBP and the potential applications by the SCUT-FBP5500.

The main contributions of this paper can be summarized as following:
\begin{enumerate}
  \item \textbf{Dataset.} We propose a new large-scale SCUT-FBP5500 benchmark dataset that has totally 5500 frontal faces with diverse properties and diverse labels, which allows construction of FBP models with different paradigms.
  \item \textbf{Benchmark Analysis.} We analyze the samples, score labels, labelers and facial landmarks of the SCUT-FBP5500 statistically, and the visualization of the data illustrates the properties of the SCUT-FBP5500.
  \item \textbf{Facial Beauty Prediction Evaluation.} Both shallow prediction model with hand-crafted feature and deep learning models are trained on the SCUT-FBP5500 for evaluation, and the results indicates the improvement of FBP based on the proposed dataset with better diversity.
\end{enumerate}



\section{Construction of SCUT-FBP5500 Dataset}
\subsection{Face Images Collection}
The SCUT-FBP5500 Dataset contains 5500 frontal, un-occluded faces aged from 15 to 60 with neutral expression. It can be divided into four subsets with different races and gender, including 2000 Asian females, 2000 Asian males, 750 Caucasian females and 750 Caucasian males. Most of the images of the SCUT-FBP5500 were collected from Internet, where some portions of Asian faces were from the DataTang~\cite{datatang} and some Caucasian faces were from the 10k US Adult database~\cite{10kUS}.

\subsection{Facial Beauty Scores and Facial Landmarks}

\begin{table}[!t]
\centering
\caption{The Outlier Number and Portion of Beauty Scores of Caucasian female (CF), Caucasian male (CM), Asian female (AF) and Asian male (AM).}
\label{tab_filer}
\begin{tabular}{|c||c|c|c|c|}
\hline
Subset & CF & CM & AF & AM \\
\hline
\hline
Total Score Num. & 45000 & 45000 & 120000 & 120000 \\
\hline
Outlier Num. & 143 & 181 & 356 & 497 \\
\hline
Outlier Portion & $0.3\%$ & $0.4\%$ & $0.3\%$ &$0.4\%$\\
\hline
\end{tabular}
\end{table}

\begin{figure}[!t]
\centering
\subfigure[Score distribution of CF]{\includegraphics[width=1.5in]{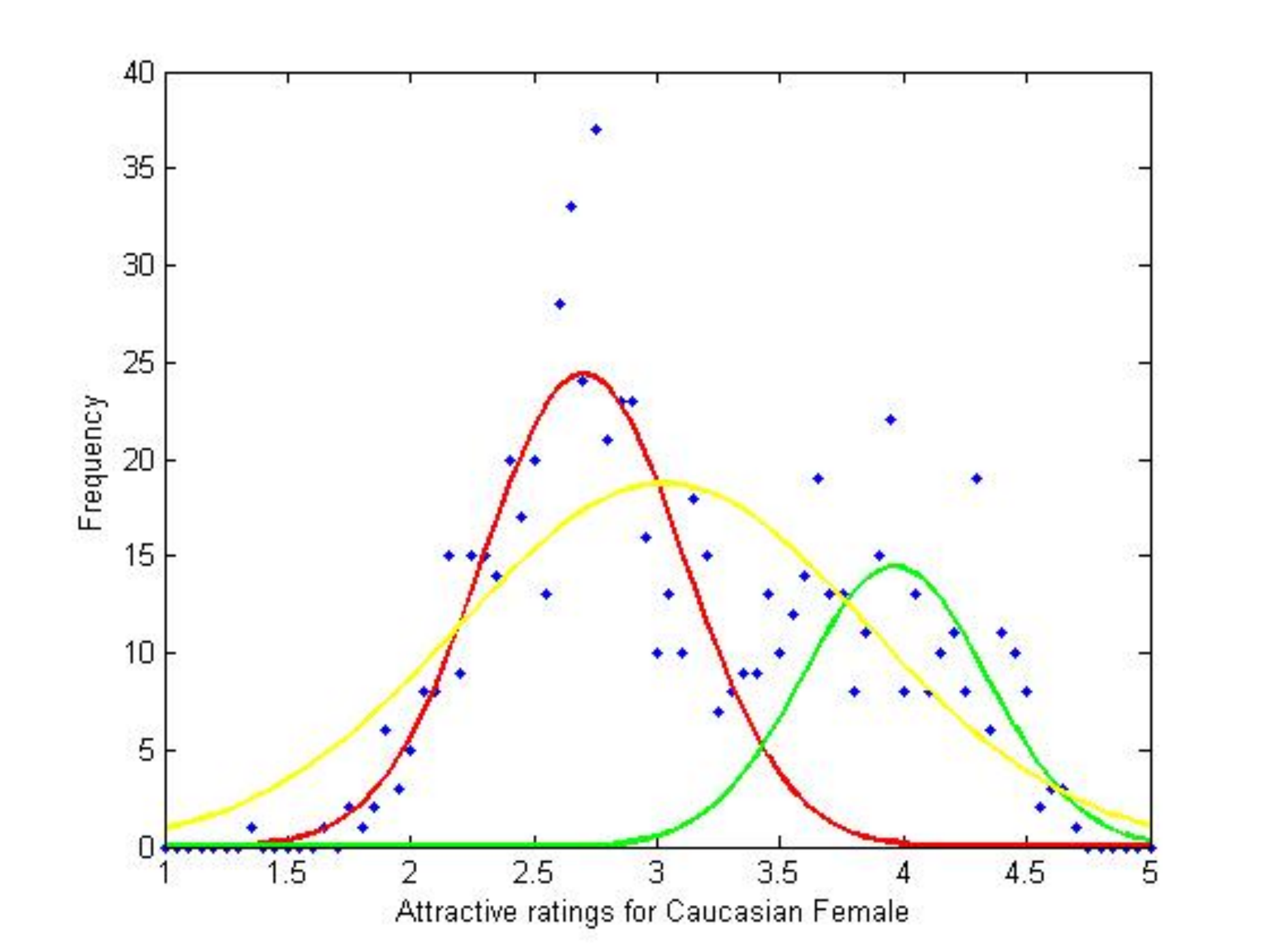}}
\subfigure[Score distribution of CM]{\includegraphics[width=1.5in]{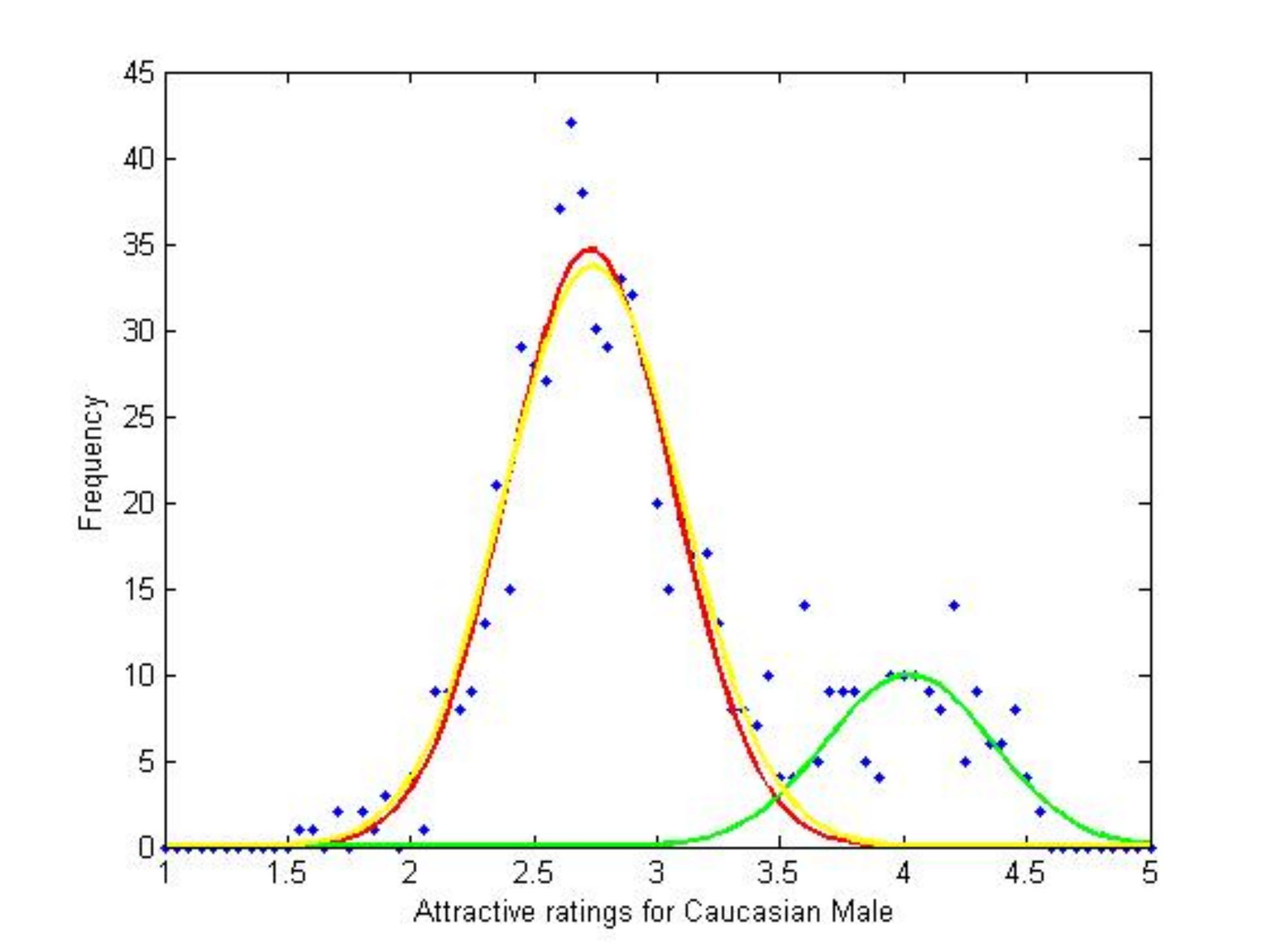}}

\subfigure[Score distribution of AF]{\includegraphics[width=1.5in]{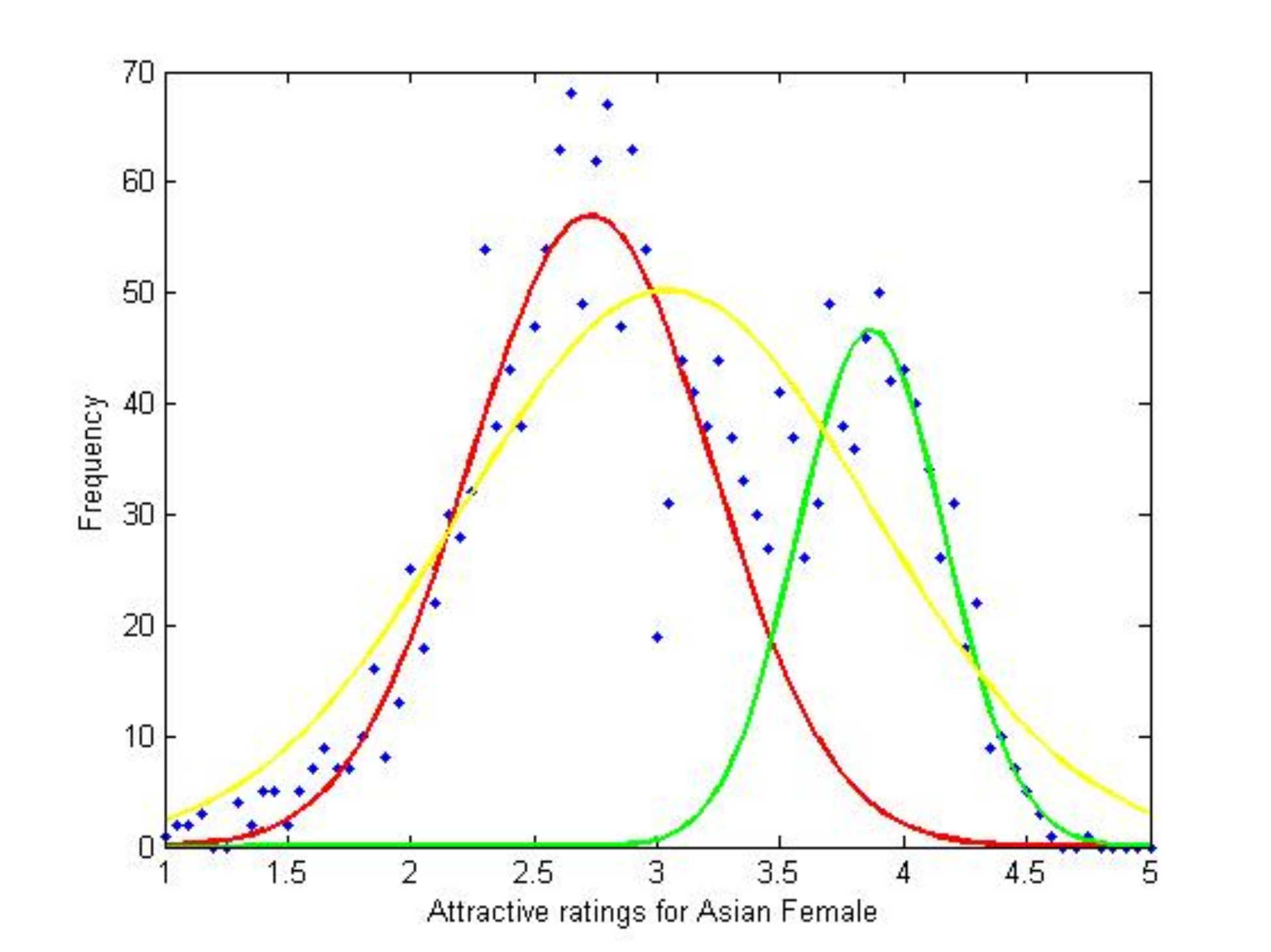}}
\subfigure[Score distribution of AM]{\includegraphics[width=1.5in]{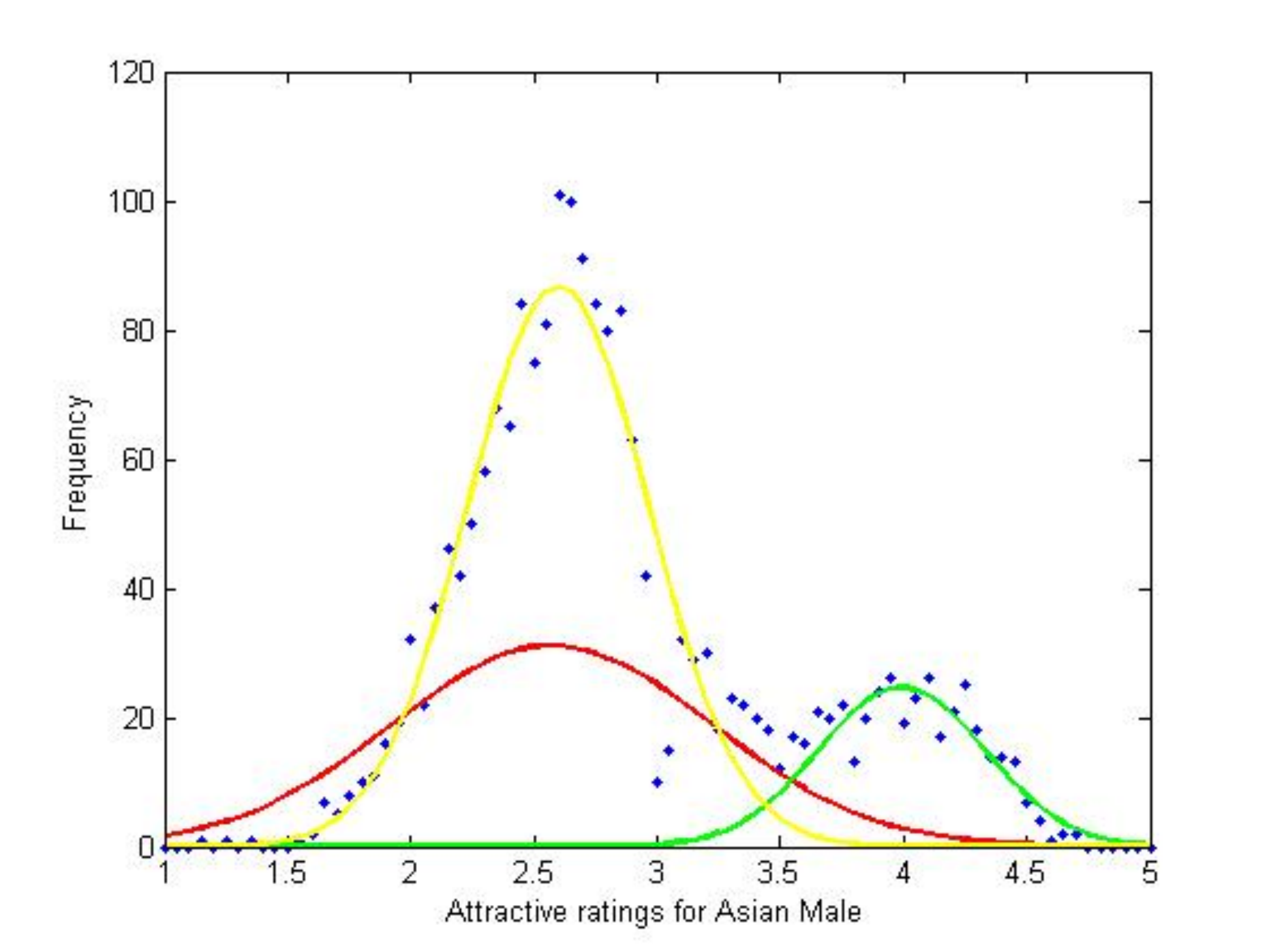}}
\caption{Gaussian fitting (yellow curve) and piecewise fitting (red and blue curve) for the visualization of the beauty score distribution of Caucasian female (CF), Caucasian male (CM), Asian female (AF) and Asian male (AM).}
\label{fig_fitting_7}
\end{figure}

All the images are labeled with beauty scores ranging from [1,~5] by totally 60 volunteers aged from 18-27 (average 21.6), where the beauty score 5 means most attractive and so on. We developed a web-based GUI system to obtain the facial beauty scores. The labeling system was deployed on the Ali Cloud, and the labeling tasks are distributed to each volunteer in crowdsourcing manners. The four subset, Asian male/female and Caucasian male/female, are labeled separately, where each face of the subset are randomly shown to the volunteer. Then, the volunteer are asked to select the beauty scores within [1,~5] for the face. To reduce the variance in the labeling process, about $10\%$ faces recurred randomly. If the correlation coefficient of the two beauty score of the same faces is less than 0.7, the volunteer would be asked to rate this face once more to decide the final score.

To allow geometric analysis of facial beauty, 86 facial landmarks are located to the significant facial components of each images, such as eyes, eyebrows, nose and mouth. A GUI landmarks location system is developed, where the original location of the landmarks are initialized by the active shape model (ASM) trained by the SCUT-FBP dataset. Then, the detected landmarks by ASM are modified manually by volunteers to ensure the accuracy.




\begin{figure}[!t]
\centering
\includegraphics[width=3in]{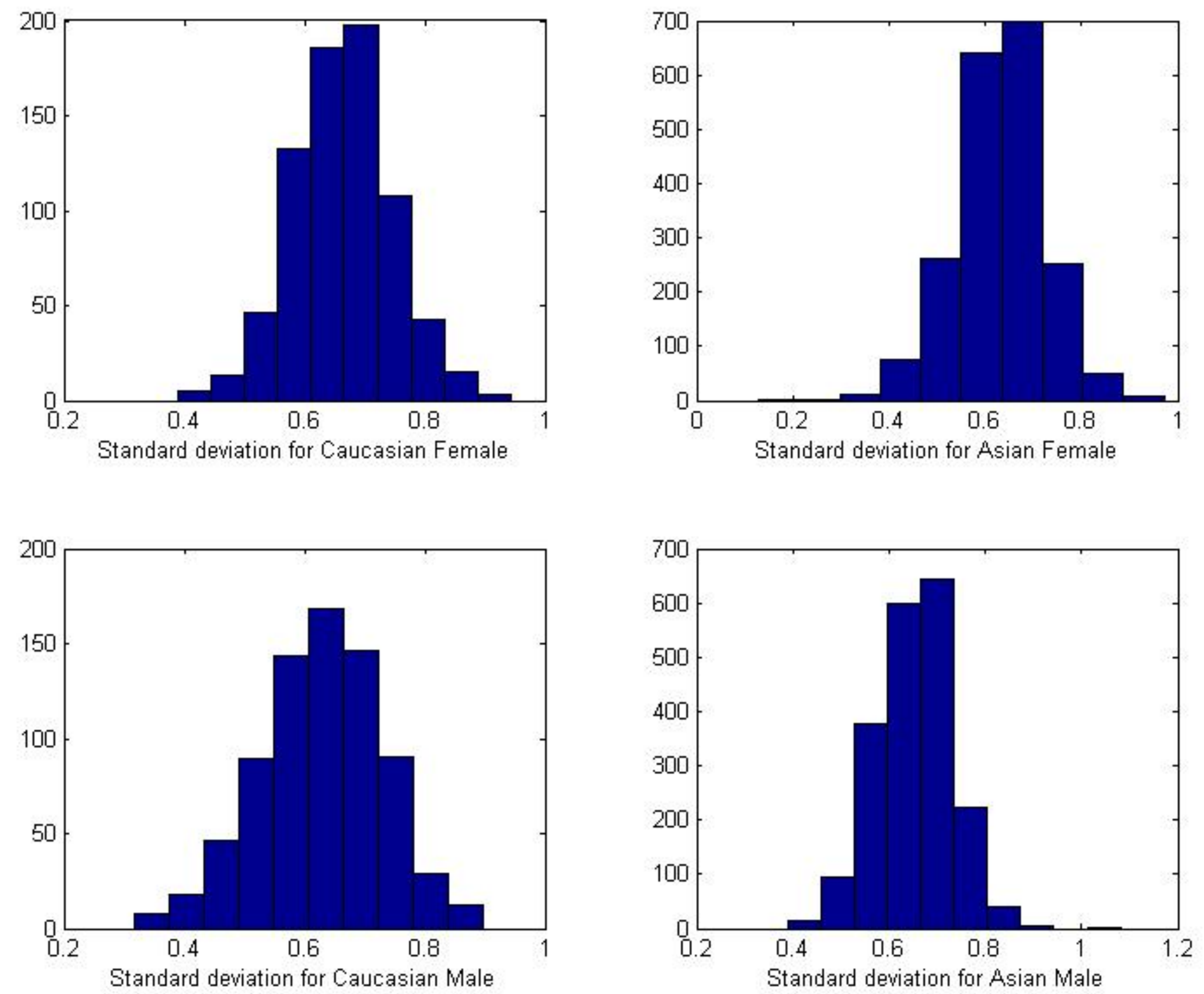}
\caption{Distribution of standard deviations of Caucasian female, Asian female, Caucasian male and Asian male, respectively.}
\label{fig_deviation}
\end{figure}

\begin{figure}[!t]
\centering
\subfigure[Box figure of CF]{\includegraphics[width=1.7in]{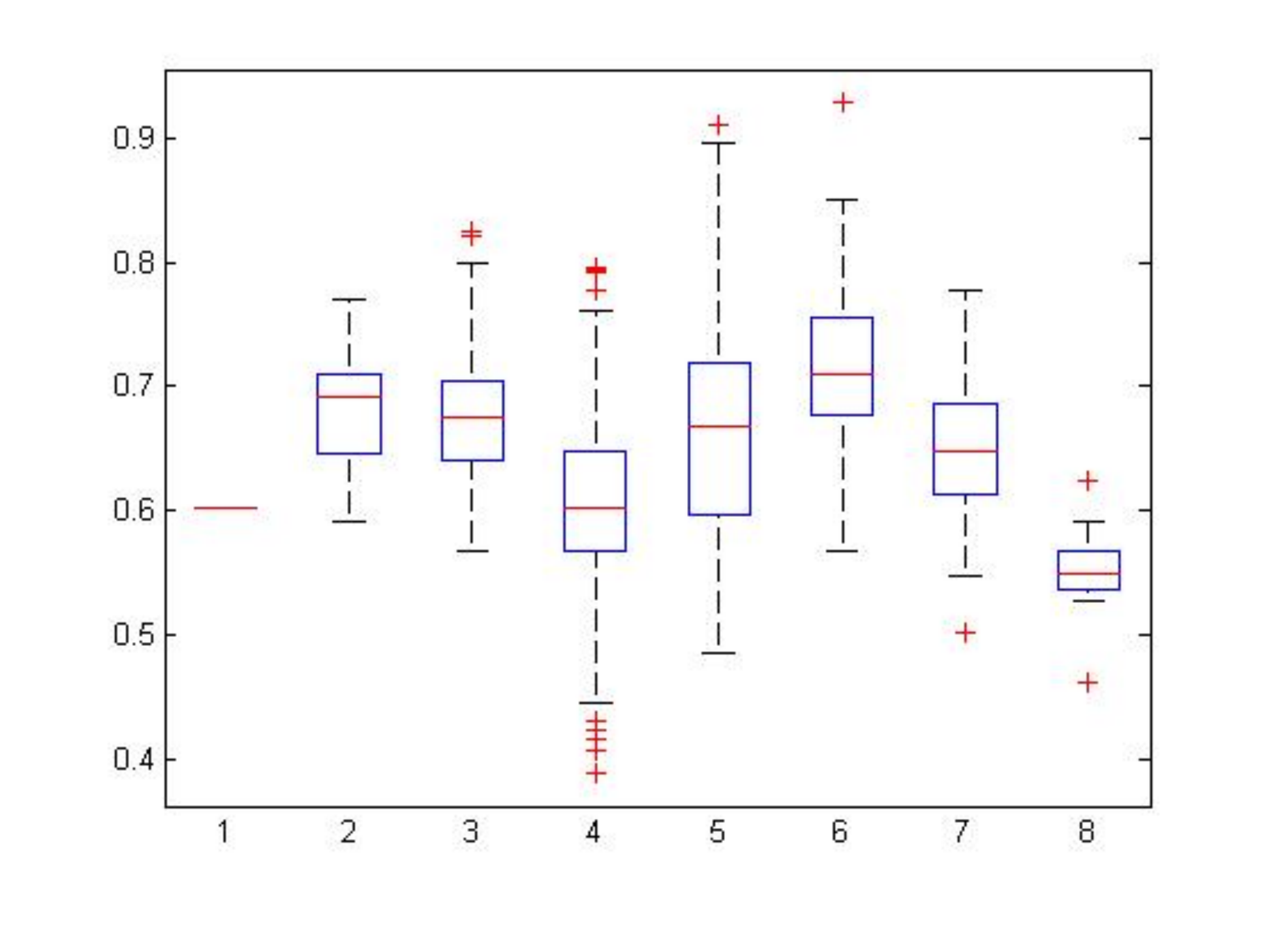}}
\subfigure[Box figure of AF]{\includegraphics[width=1.7in]{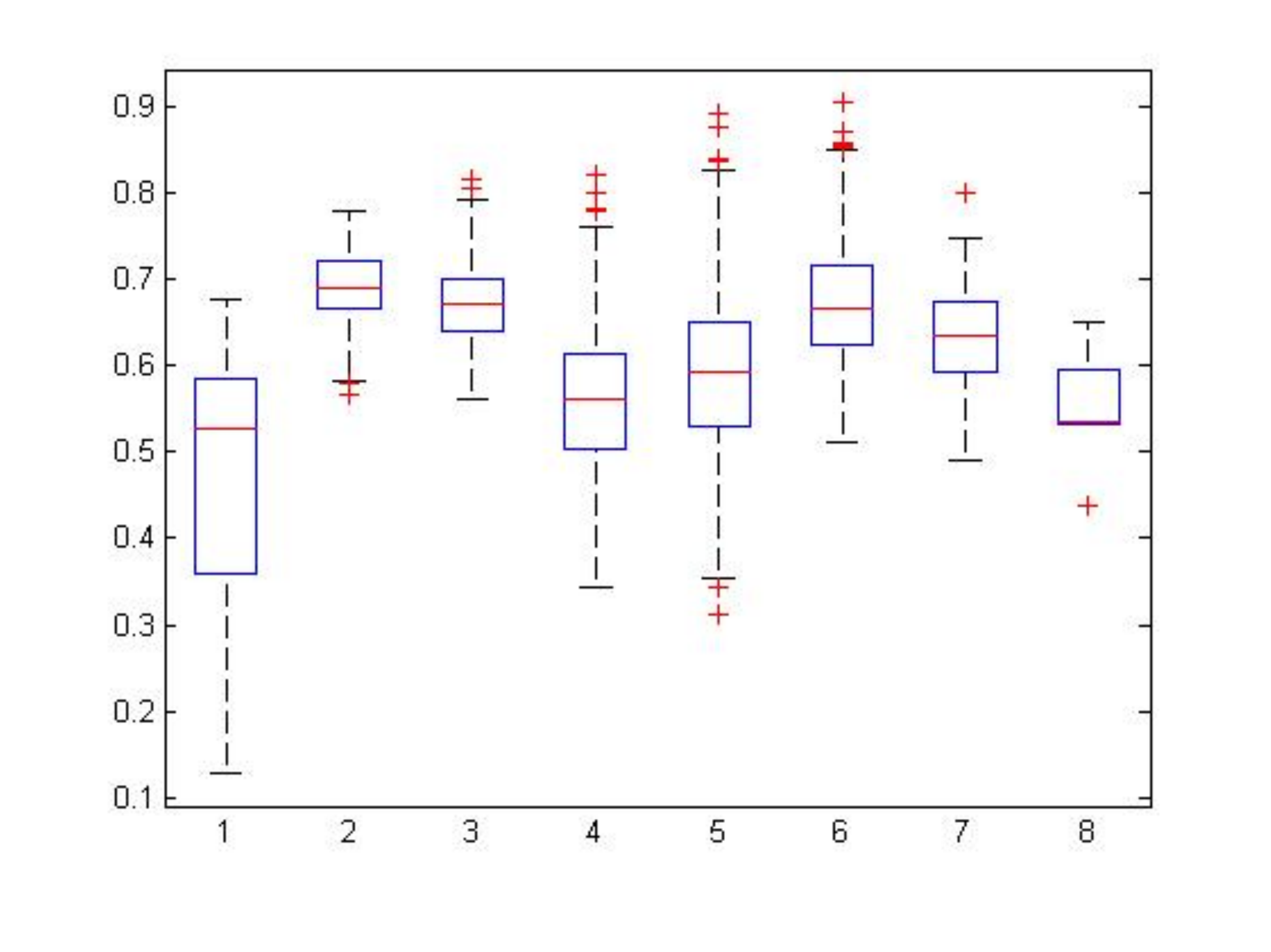}}

\subfigure[Box figure of CM]{\includegraphics[width=1.7in]{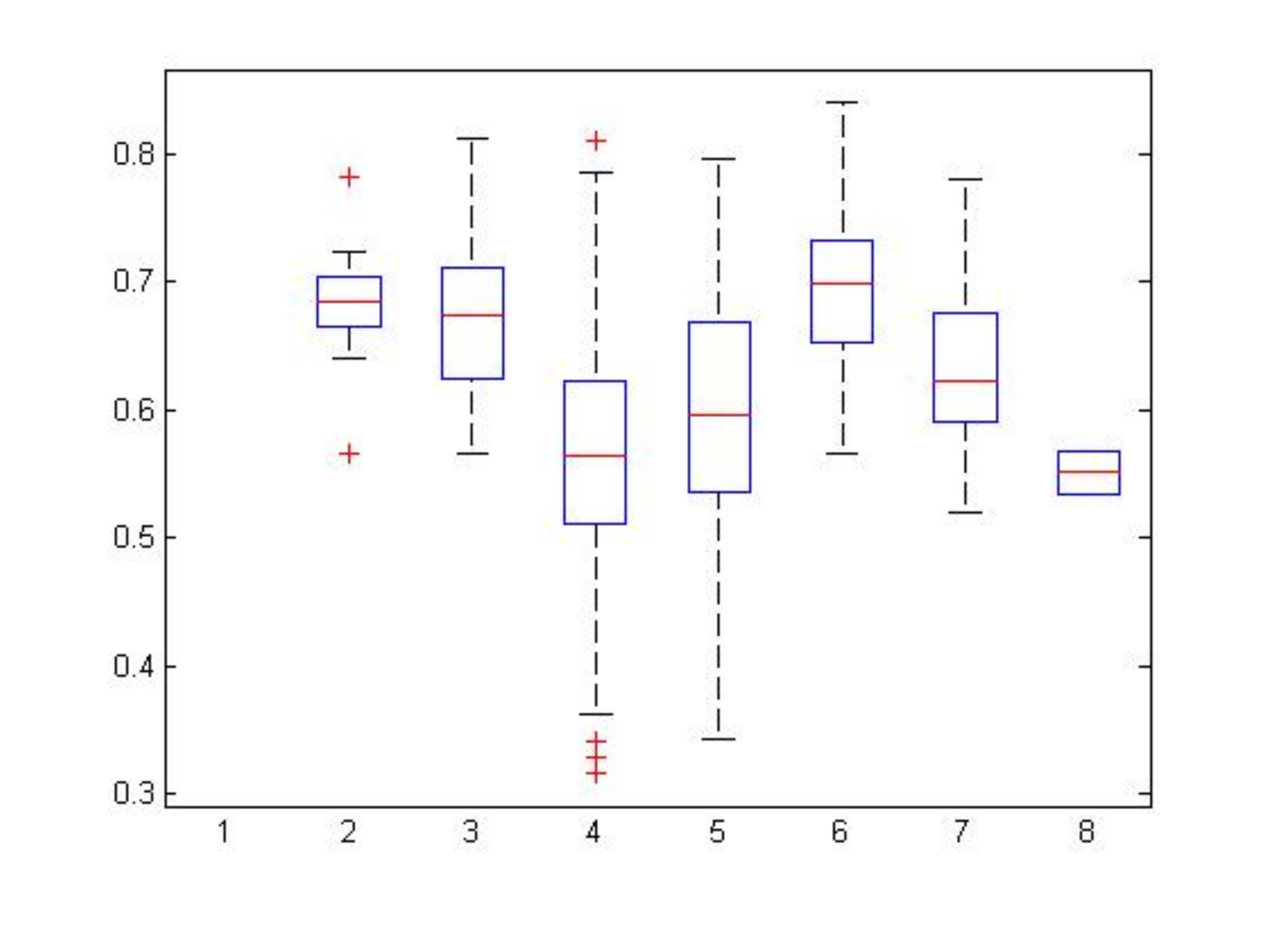}}
\subfigure[Box figure of AM]{\includegraphics[width=1.7in]{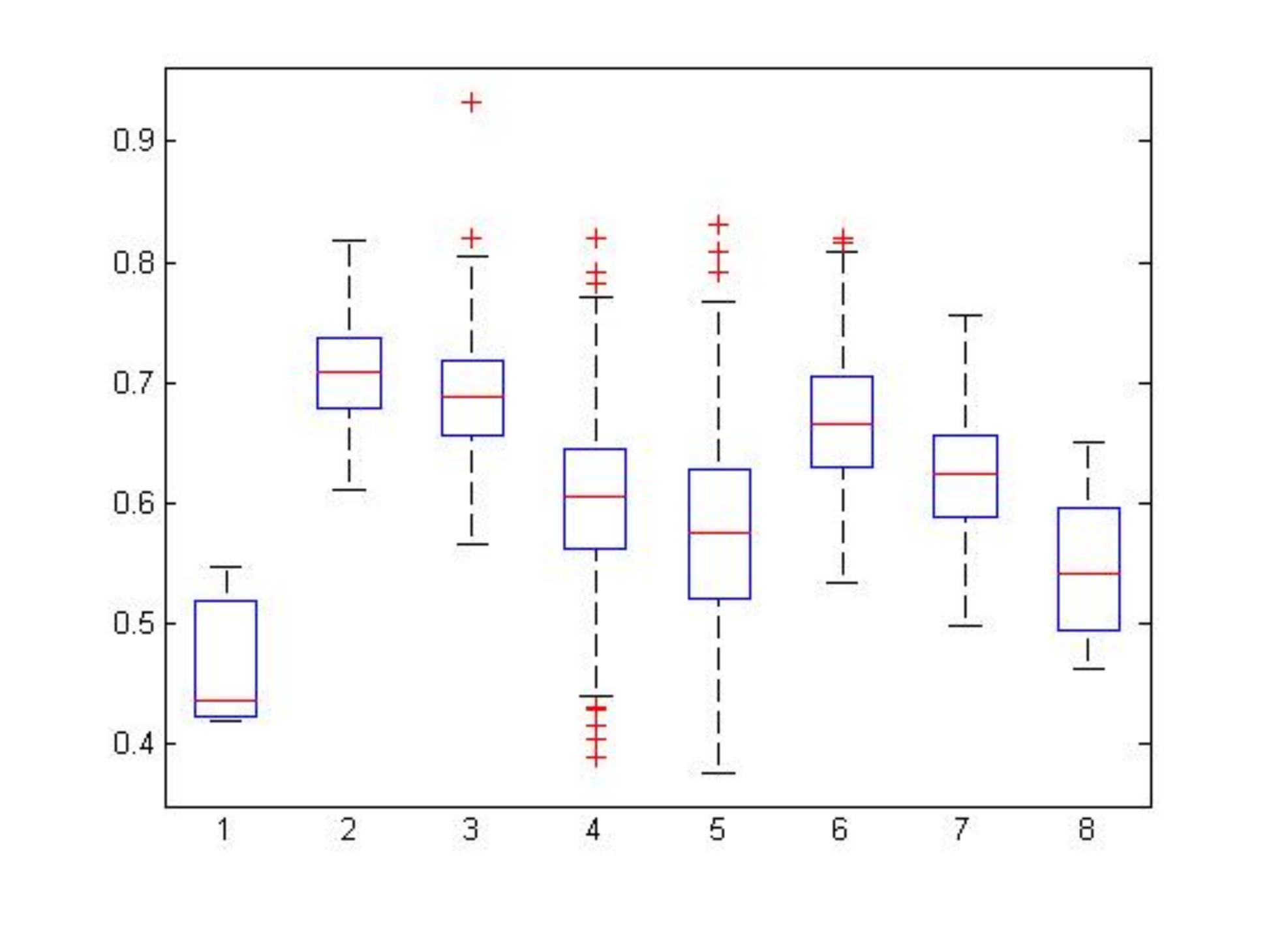}}
\caption{Box figures of standard deviations of Caucasian female (CF), Asian female (AF), Caucasian male (CM) and Asian male (AM), respectively.}
\label{fig_deviation2}
\end{figure}

\section{Benchmark Analysis of the SCUT-FBP5500}
We made benchmark analysis of the beauty scores, labelers and face landmarks of the SCUT-FBP5500 with different gender and races, including Asian female (AF), Asian male (AM), Caucasian female (CF) and Caucasian male (CM).

\begin{table}[t]
\centering
\caption{Correlation Coefficients between Male and Female Labelers for Beauty Score of Caucasian female (CF), Caucasian male (CM), Asian female (AF) and Asian male (AM).}
\label{tab_gender}
\begin{tabular}{|c||c|c|c|c|c|}
\hline
 & CF & AF & CM & AM & All Faces\\
\hline
Female Labelers & 0.785 & 0.800 & 0.747 & 0.793 & 0.785 \\
\hline
Male Labelers & 0.791 & 0.795 & 0.763 & 0.797 & 0.781 \\
\hline
All Labelers & 0.788 & 0.785 & 0.743 & 0.782 & 0.770 \\
\hline
\end{tabular}
\end{table}

\begin{figure}[t]
\centering
\includegraphics[width=2.7in]{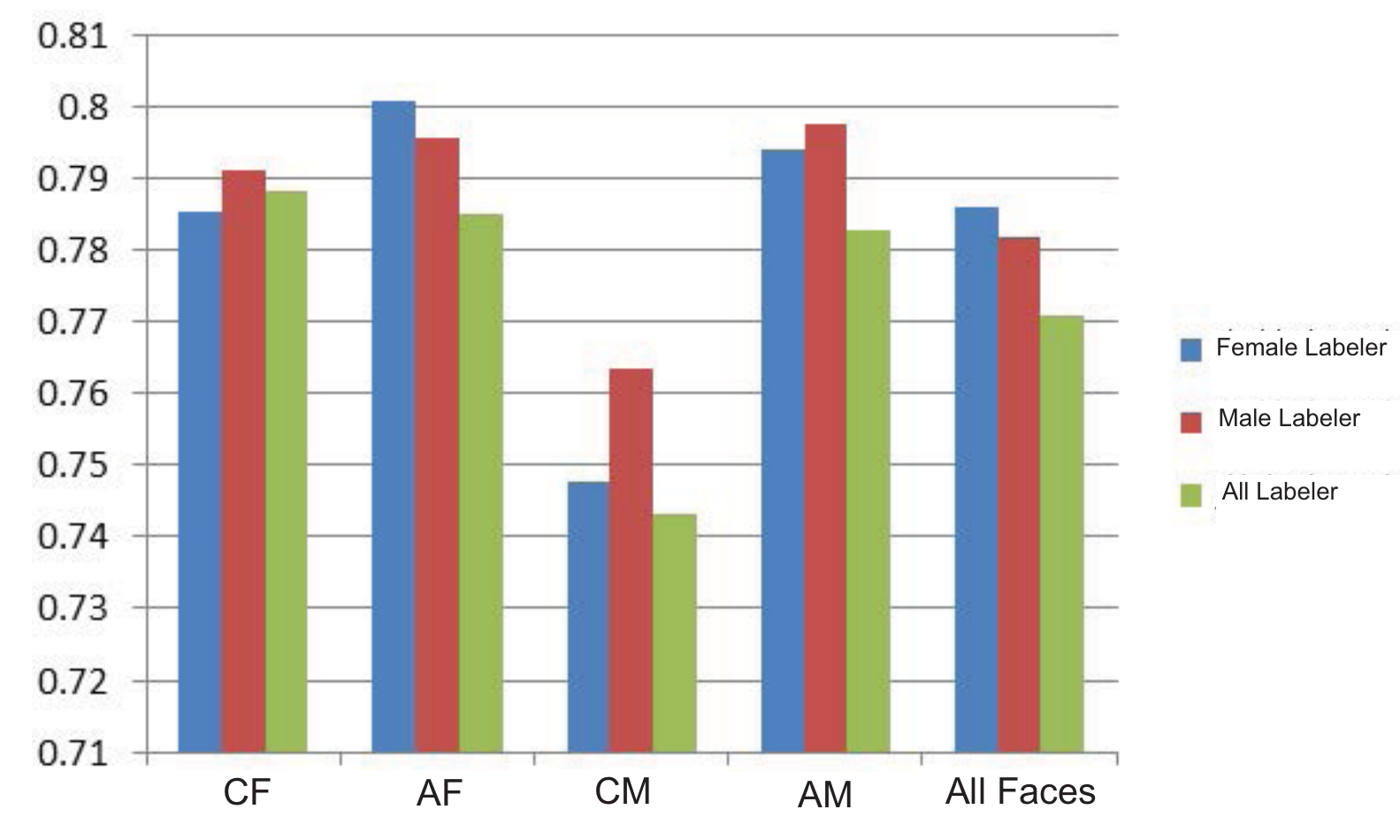}
\caption{Correlation Coefficient of male and female labelers for beauty score of CF, AF, CM, AM and all the faces in SCUT-FBP5500.}
\label{fig_gender}
\end{figure}

\begin{figure}[!t]
\centering
\subfigure[PCA Analysis of AF]{\includegraphics[width=3in]{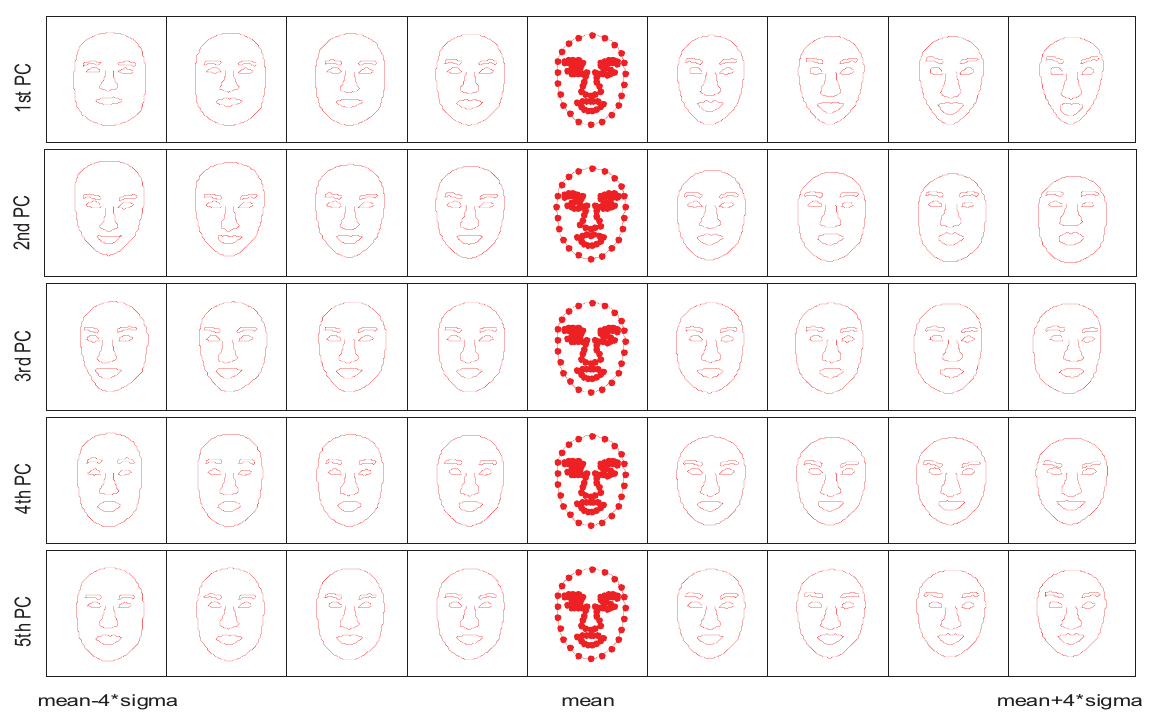}}

\subfigure[PCA Analysis of AM]{\includegraphics[width=3in]{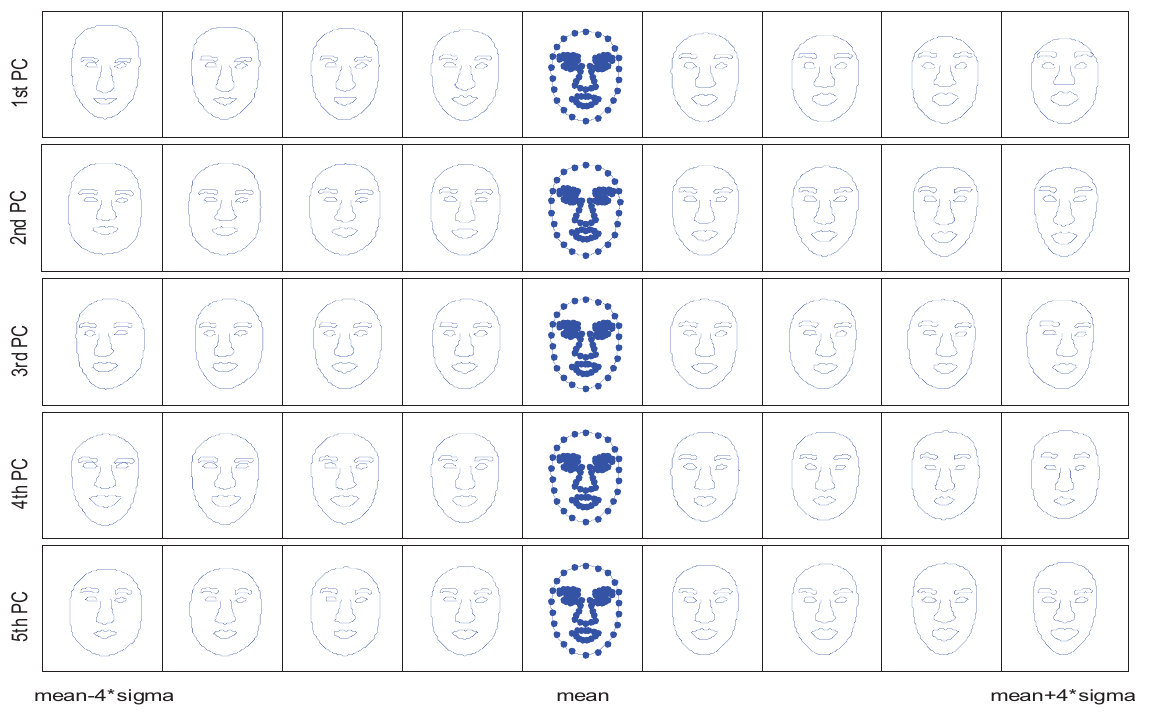}}
\caption{PCA analysis of face landmarks of Asian female (AF) and Asian male (AM).}
\label{fig_pca}
\end{figure}

\subsection{Distribution of Beauty Scores}
We visualize the distribution of the beauty scores of AF, AM, CF and CM, respectively. To obtain better visualization, we preprocess the data and filter the outliers of the beauty scores. We regard the average score of all the 60 labelers as the ground-truth. If the score of specific labeler for the same face differs from the ground-truth over 2, the score is treated as outlier and is removed from the distribution visualization. The number and portion of the outliers in the four subset are listed in Table~\ref{tab_filer}, and the small portion of outlier indicate the reliability of the labeling process for beauty score.

Since the outlier portion of the beauty score is tiny, the distributions of the original data and the preprocessed data is mostly similar. Therefore, we visualized the score distribution of the SCUT-FBP5500 using the preprocessed data for all the four subset, respectively. Two distribution fitting schemes are used: one is Gaussian fitting (yellow curve), the other is piecewise fitting (red and blue curve), as shown in Fig.~\ref{fig_fitting_7}. The results indicates that the beauty scores of all the four subset can be approximately fitted by a mixed distribution model with two Gaussian components.




\subsection{Standard Deviation of Beauty Scores}
We calculate the standard deviation of the scores gathered from different labelers to the ground-truth, and illustrate the results as histogram in Fig.~\ref{fig_deviation} and as box figure in Fig.~\ref{fig_deviation2}. We observe that the distribution of standard deviations is similar to Gaussian distribution, and most of the standard deviations are within a reasonable range of [0.6,~0.7].

\subsection{Correlation of Male/Female Labelers}
In this subsection, we investigate the correlation between the male and female Asian labelers for the beauty scores, as shown in Table~\ref{tab_gender} and Fig.~\ref{fig_gender}. We observe that the correlation of Asian faces is persistently larger than Caucasian. It is consistent to the psychological research that human have better facial beauty perception for the faces from the same race.

\subsection{PCA Analysis of Facial Geometry}
We visualize the 86-points face landmarks of the dataset using principle component analysis (PCA). Fig.~\ref{fig_pca} illustrate the mean and the five first principle component of the facial geometry of Asian female (AF) and Asian male (AM), where the landmarks data of Caucasian share similar distribution to Asian faces. We observe that the face shape is one of the main component influence the face geometry of beauty, which is consistent to related psychological research and previous works in~\cite{2011zhang,4}

\section{FBP Evaluation via Hand-Crafted Feature and Shallow Predictor}
This section, we evaluate the SCUT-FBP5500 using the hand-crafted feature with shallow predictor, while the next section introduce some state-of-the-art deep learning model to achieve FBP.

\subsection{Geometric Feature with Shallow Predictor}
We extract a 18-dimensional ratio feature vector from the faces and formulate FBP based on different regression models, such as the linear regression (LR), Gaussian regression (GR), and support vector regression (SVR). Comparison were performed for Caucasian female/male and Asian female/male subsets, and the performance of different model are measured using pearson correlation coefficient (PC)~\cite{2006nc_Dror}, maximum absolute error (MAE) and root mean square error (RMSE) after 10 folds cross validation. The results are listed in Table~\ref{tab_geometric} and Table~\ref{tab_geometric_all}, which can be regarded as a baseline for the geometric analysis of FBP.


\begin{table}[!ht]
\centering
\caption{Facial beauty prediction using geometric feature with shallow models for subsets of different races and gender}
\label{tab_geometric}
\begin{tabular}{c|ccc|ccc}
\hline
\multirow{2}{*}{} & \multicolumn{3}{c|}{Asian Female} & \multicolumn{3}{c}{Asian Male}\\
\hline
 & LR & GR & SVR & LR & GR & SVR  \\
\hline
PC & 0.6771 & 0.7057 & 0.7008 & 0.6348 & 0.6923 & 0.6816  \\
MAE & 0.402 & 0.387 & 0.3876 & 0.3894 & 0.3572 & 0.356  \\
RMSE & 0.5246 & 0.5057 & 0.5089 & 0.5085 & 0.4752 & 0.4823 \\
\hline
\hline
\multirow{2}{*}{}  & \multicolumn{3}{c|}{Caucasian Female} & \multicolumn{3}{c}{Caucasian Male}\\
\hline
 & LR & GR & SVR & LR & GR & SVR \\
\hline
PC &  0.6809 & 0.7263 & 0.7093 & 0.6063 & 0.63 & 0.6397 \\
MAE &  0.3986 & 0.3862 & 0.4001 & 0.3871 & 0.3689 & 0.3617 \\
RMSE  & 0.5239 & 0.4908 & 0.5087 & 0.4899 & 0.4784 & 0.4739 \\
\hline
\end{tabular}
\end{table}

\begin{table}[!ht]
\centering
\caption{Facial beauty prediction using Geometric Feature with shallow models for the whole dataset}
\label{tab_geometric_all}
\begin{tabular}{cccc}
\toprule
 & Linear Regression & Gaussian Regression & SVR \\
\midrule
PC & 0.5948 & 0.6738 & 0.6668 \\
MAE & 0.4289 & 0.3914 & 0.3898 \\
RMSE & 0.5531 & 0.5085 & 0.5132 \\
\bottomrule
\end{tabular}
\end{table}

\begin{table}[!t]
\centering
\caption{Facial beauty prediction using Gabor feature with two sampling scheme on whole dataset}
\label{tab_gabor_all}
\begin{tabular}{c|cc|cc}
\hline
\multirow{2}{*}{} & \multicolumn{2}{c|}{86-keypoints} & \multicolumn{2}{c}{64UniSample} \\
\hline
\hline
 & GR & SVR & GR & SVR  \\
\hline
PC & 0.7472 & 0.6691 & 0.6764 &0.8065 \\
MAE & 0.3554 & 0.3891 & 0.4014 & 0.3976 \\
RMSE & 0.4599 & 0.5065 & 0.5177 & 0.5126  \\
\hline
\end{tabular}
\end{table}

\begin{table}[t]
\caption{Comparison of AlexNet~\cite{krizhevsky2012imagenet}, ResNet-18~\cite{he2016deep} and ResNeXt-50~\cite{xie2016aggregated} in measurement of PC, MAE and RMSE by 5-fold cross validation}
\label{table_deep1}
\centering
\begin{tabular}{c|c|c|c|c|c|c}
\hline
\hline
\textbf{PC} & 1 & 2 & 3 & 4 & 5 & Average \\
\hline
AlexNet  & 0.8667 & 0.8645 & 0.8615 & 0.8678 & 0.8566 & 0.8634 \\
ResNet-18 & 0.8847 & 0.8792 & 0.8929 & 0.8932 & 0.9004 & 0.89 \\
ResNeXt-50 & \textbf{0.8985} & \textbf{0.8932} & \textbf{0.9016} & \textbf{0.899} & \textbf{0.9064} & \textbf{0.8997} \\
\hline
\hline
\textbf{MAE} & 1 & 2 & 3 & 4 & 5 & Average \\
\hline
AlexNet & 0.2633 & 0.2605 & 0.2681 & 0.2609 & 0.2728 & 0.2651 \\
ResNet-18 & 0.248 & 0.2459 & 0.243 & 0.2383 & 0.2383 & 0.2419 \\
ResNeXt-50 & \textbf{0.2306} & \textbf{0.2285} & \textbf{0.226} & \textbf{0.2349} & \textbf{0.2258} & \textbf{0.2291} \\
\hline
\hline
\textbf{RMSE} & 1 & 2 & 3 & 4 & 5 & Average \\
\hline
AlexNet & 0.3408 & 0.3449 & 0.3538 & 0.3438 & 0.3576 & 0.3481 \\
ResNet-18 & 0.3258 & 0.3286 & 0.3184 & 0.3107 & 0.2994 & 0.3166 \\
ResNeXt-50 & \textbf{0.3025} & \textbf{0.3084} & \textbf{0.3016} & \textbf{0.3044} & \textbf{0.2918} & \textbf{0.3017} \\
\hline
\end{tabular}
\end{table}

\begin{table}[!ht]
\centering
\caption{Comparison of AlexNet~\cite{krizhevsky2012imagenet}, ResNet-18~\cite{he2016deep} and ResNeXt-50~\cite{xie2016aggregated} in measurement of PC, MAE and RMSE by $60\%$ training and $40\%$ testing}
\label{table_deep2}
\begin{tabular}{cccc}
\toprule
 & AlexNet & ResNet-18 & ResNeXt-50 \\
\midrule
PC & 0.8298 &	0.8513	& \textbf{0.8777} \\
MAE & 0.2938 &	0.2818	& \textbf{0.2518}\\
RMSE & 0.3819 &	0.3703	& \textbf{0.3325} \\
\bottomrule
\end{tabular}
\end{table}

\subsection{Appearance Feature with Shallow Predictor}
\begin{figure}[!t]
\centering
\includegraphics[width=2in]{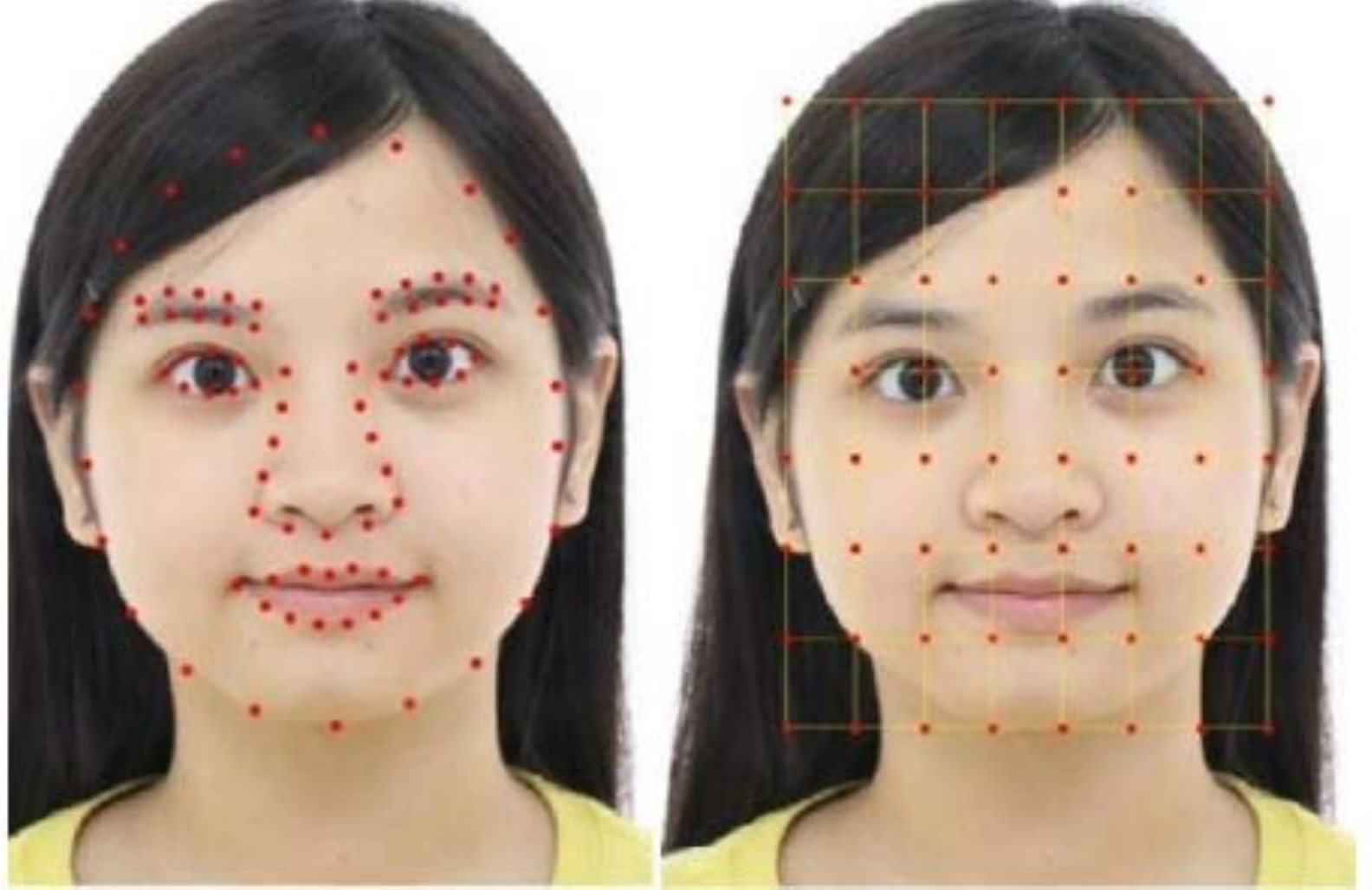}
\caption{Two sampling schemes to extract the appearance feature of FBP, where the left one is the 86-keypoints method and the right one is the UniSamplePoint method.}
\label{fig_gabor_sample}
\end{figure}
We extract 40 Gabor feature maps from every original image in five directions and eight angles. Then, we obtain the appearance feature of FBP using two different sampling schemes that extracts some component of the Gabor feature maps as following:
\begin{itemize}
\item Sample 86-keypoints from each 40 Gabor feature maps to obtain a 3340-dimensional feature vector, as shown in the right sub-figure of Fig.~\ref{fig_gabor_sample};
\item Use 64UniSample to obtain a 2560-dimensional feature vector, as shown in the left sub-figure of Fig.~\ref{fig_gabor_sample}.
\end{itemize} Finally, we use PCA to reduce the extracted feature dimension before we train the predictor. The results of the appearance-based shallow predictors for all the data are shown in Table~\ref{tab_gabor_all}.


\section{FBP Evaluation via Deep Predictor}
We evaluate three recently proposed CNN models with different structures for FBP, including AlexNet~\cite{krizhevsky2012imagenet}, ResNet-18~\cite{he2016deep} and ResNeXt-50~\cite{xie2016aggregated}. All these CNN models are trained by initializing weights using networks pre-trained on the ImageNet dataset. The evaluation of were performed under two different experiment settings as following:
\begin{enumerate}
  \item The models were trained and tested using 5-fold cross validation, which means each fold containing $20\%$ samples (1100 images). The accuracy of each fold and the average of all the 5 fold are shown in Table~\ref{table_deep1}.
  \item The models were trained using $60\%$ samples (3300 images), and tested with the rest $40\%$ (2200 images). The results are shown in Table~\ref{table_deep2}.
\end{enumerate}

The results illustrates that the deepest CNN-based ResNeXt-50 model~\cite{xie2016aggregated} obtains the best performance comparing to the ResNet-18 and AlexNet in both the experiment setting. It can be observed that all the deep CNN model are superior to the shallow predictor with hand-crafted geometric feature in Table~\ref{tab_geometric_all} or appearance feature in Table~\ref{tab_gabor_all}. It indicates the effectiveness and powerfulness of the end-to-end feature learning deep model for FBP.

Comparing the results of Table~\ref{table_deep1} and Table~\ref{table_deep2}, we also find that the accuracy of all the 5-fold cross validation is slightly higher than the results of the split of $60\%$ training and $40\%$ testing. One of the reasons may be due to the amounts and diversity of the training samples, since the 5-fold cross validation use $80\%$ samples to train the models. This observation indicates that the data augmentation techniques may further improve the performance of the deep FBP model, which merits exploring in the future.

\section{Conclusion}
In this paper, we introduce a new diverse benchmark dataset, called SCUT-FBP5500, to achieve multi-paradigm facial beauty prediction. The SCUT-FBP5500 dataset has totally 5500 frontal faces with diverse properties (male/female, Asian/Caucasian, ages) and diverse labels (face landmarks, beauty scores within [1,~5], beauty score distribution). Benchmark analysis have been made for the beauty scores and landmarks in SCUT-FBP5500, and the visualization of the data shows the statistical properties of the dataset. Since the SCUT-FBP5500 is designed for multi-paradigm, it can be adapted to different FBP models for different tasks, like appearance-based or shape-based model for facial beauty classification/regression/ranking. We evaluated the SCUT-FBP5500 using different combinations of feature and predictor, and deep learning models, where the results indicates the reliability of the dataset.






%

\balance

\end{document}